\documentclass[conference]{IEEEtran}
\IEEEoverridecommandlockouts
\usepackage{cite}
\usepackage{amsmath,amssymb,amsfonts}
\usepackage{algorithmic}
\usepackage{graphicx}
\usepackage{textcomp}
\usepackage{xcolor}
\def\BibTeX{{\rm B\kern-.05em{\sc i\kern-.025em b}\kern-.08em
    T\kern-.1667em\lower.7ex\hbox{E}\kern-.125emX}}

\newcommand \ignore[1]{}

\usepackage{url}
\usepackage{float}
\usepackage{colortbl}

\usepackage{caption}
\usepackage{subcaption}

\begin{document}

\title{Cough Classification using Few-Shot Learning}

\author{
\IEEEauthorblockN{
    Yoga Disha Sendhil Kumar, 
    Manas V Shetty, 
    Sudip Vhaduri
}
\IEEEauthorblockA{
    Department of Computer and Information Technology, Purdue University, \\
    West Lafayette, IN, USA \\
    \{ysendhil, shetty76, svhaduri\}@purdue.edu
}
}

\maketitle

\begin{abstract}
This paper investigates the effectiveness of few-shot learning for respiratory sound classification, focusing on cough-based detection of COVID-19, Flu, and healthy conditions. We leverage Prototypical Networks with spectrogram representations of cough sounds to address the challenge of limited labeled data.
Our study evaluates whether few-shot learning can enable models to achieve performance comparable to traditional deep learning approaches while using significantly fewer training samples. Additionally, we compare multi-class and binary classification models to assess whether multi-class models can perform comparably to their binary counterparts.
Experimental findings show that few-shot learning models can achieve competitive accuracy. Our model attains 74.87\% accuracy in multi-class classification with only 15 support examples per class, while binary classification achieves over 70\% accuracy across all class pairs. Class-wise analysis reveals Flu as the most distinguishable class, and Healthy as the most challenging. Statistical tests (paired $t$-test $p=0.149$, Wilcoxon $p=0.125$) indicate no significant performance difference between binary and multi-class models, supporting the viability of multi-class classification in this setting. These results highlight the feasibility of applying few-shot learning in medical diagnostics, particularly when large labeled datasets are unavailable.
\end{abstract}


\section{Introduction}

\subsection{Background and Motivation}

Respiratory sound classification has gained significant attention due to its potential applications in disease diagnosis, remote healthcare, and early screening for respiratory illnesses. The COVID-19 pandemic highlighted the need for rapid, accessible, and non-invasive diagnostic tools, with research showing that cough sound analysis could provide valuable insights into disease differentiation \cite{b1}.

Deep learning models have demonstrated high accuracy in medical sound classification, but they often require large-scale labeled datasets to generalize well. In particular, convolutional neural networks (CNNs) and transformer-based architectures have shown promise in medical sound classification tasks \cite{b2}. However, collecting and annotating such datasets, especially for rare diseases, is challenging.

To explore these challenges in practice, we develop a custom cough classification pipeline using spectrogram representations of audio signals derived from datasets such as COUGHVID, Coswara, and FluSense. Leveraging PyTorch and Librosa, we preprocess cough audio recordings into Mel-spectrograms, which are then used as input to CNN-based few-shot models. This approach allows us to replicate the challenges of real-world deployment, where large-scale annotated datasets are not always available.

Our work emphasizes episodic training strategies to simulate low-resource diagnostic settings and adopts Prototypical Networks, which have shown promise in healthcare audio classification despite limited sample availability.

Recent studies have also highlighted the application of transfer learning in smart systems, such as disease monitoring using smartphones \cite{vhaduri2023transfer}, with promising opportunities in other areas including risk understanding \cite{vhaduri2022predicting,vhaduri2023bag,vhaduri2024mwiotauth}, user authentication through physiological data \cite{vhaduri2023implicit,vhaduri2017towards,vhaduri2018biometric,dibbo2022phone,cheung2020continuous,cheung2020context}, tracking well-being \cite{vhaduri2022understanding,vhaduri2015design}, evaluating sleep quality \cite{chen2020estimating,vhaduri2020nocturnal}, monitoring physical health and respiratory symptoms \cite{vhaduri2022understanding,vhaduri2020adherence,dibbo2021effect,vhaduri2023environment}, managing stress and improving mental health \cite{vhaduri2021deriving,dibbo2021visualizing,vhaduri2021predicting,kim2020understanding}, and discovering behavioral patterns linked to places of interest \cite{vhaduri2018opportunisticTBD,vhaduri2018opportunisticICHI}. Few-Shot Learning has also been applied to audio-based classification of insect species for ecological monitoring \cite{yogadisha2024insect}, and in facial recognition-based mood detection systems that adapt music recommendations based on user affect \cite{yogadisha2022music}.

\subsection{Contributions}

This study explores the effectiveness of few-shot learning techniques in the context of cough sound classification. Our work contributes to the growing body of research in this area through the following efforts:

\begin{itemize}

    \item We construct a tailored dataset by combining audio cough samples from three classes—Healthy, COVID-19, and Flu and convert them into Mel-spectrograms using Librosa. These spectrograms are then formatted to fit the input dimensions of ResNet-based few-shot classifiers. This custom preprocessing pipeline ensures our models are optimized for real-world clinical data variability.

    \item We enforce balanced train-test splits for each class to avoid class imbalance and ensure fair evaluation of binary versus multi-class models. The splits are designed to support few-shot learning tasks under both low-data and episodic training conditions.

    \item We integrate our dataset with the EasyFSL framework to train Prototypical Networks using episodic sampling and N-way, K-shot tasks, which align with real-world healthcare applications where labeled samples are scarce.

    \item We focus primarily on assessing whether multi-class classification models—developed using few-shot learning techniques—can perform comparably to their binary counterparts within a 10--15\% accuracy margin. This is framed as our main research hypothesis:

    \begin{itemize}
        \item H\textsubscript{0} (Null Hypothesis): No significant difference in accuracy between multi-class and binary models.
        \item H\textsubscript{a} (Alternative Hypothesis): Multi-class models perform significantly worse than their binary counterparts.
    \end{itemize}

    \item We conduct statistical hypothesis testing to quantify the significance of observed performance differences across model types and training regimes, thereby providing a rigorous basis for our conclusions.

    \item We compare the performance of our Few-Shot Learning-based models against existing state-of-the-art, big data-driven approaches from the literature. This comparison highlights the potential of Few-Shot Learning to achieve strong performance with significantly reduced training data, which is especially valuable in data-constrained healthcare settings.
\end{itemize}

\section{Related Work}

\subsection{Cough Classification}

The classification of respiratory sounds, particularly coughs, has gained significant interest as a non-invasive diagnostic tool for respiratory illnesses. Recent studies have demonstrated that deep learning models can analyze cough recordings to detect conditions such as COVID-19 and the Flu with promising accuracy \cite{am}. Several datasets, including COUGHVID \cite{b112}, Coswara \cite{b113}, and FluSense \cite{b4}, have been developed to facilitate research in this domain.

Early studies leveraged traditional signal processing techniques such as Mel-frequency cepstral coefficients (MFCCs) and spectrogram-based features to analyze cough sounds \cite{b1}. More recent work has shifted toward deep learning approaches, employing convolutional neural networks (CNNs) and recurrent neural networks (RNNs) to automate feature extraction and classification \cite{b6}. CNN-based models have demonstrated high accuracy in distinguishing COVID-19 from non-COVID coughs, outperforming classical machine learning techniques \cite{b7}. While handcrafted features were previously common in cough detection tasks, deep learning approaches using learned representations have shown superior performance in complex classification scenarios \cite{b8}.

In our work, we replicate and extend these methods by transforming raw cough recordings into Mel-spectrograms, which are known to preserve temporal and frequency-based patterns critical for disease classification. This approach aligns with previous research that emphasized the effectiveness of spectrograms and MFCCs in respiratory signal classification.

\subsection{Few-Shot Learning for Medical Applications}

Few-shot Learning (FSL) techniques have emerged as a promising solution for medical applications where labeled data is limited. Prototypical Networks \cite{b3}, a popular metric-based FSL approach, have been successfully applied to image classification and medical imaging tasks \cite{b10}. The underlying principle of Prototypical Networks is to compute class prototypes in an embedding space and classify new samples based on distance to these prototypes.

Recent research has explored the application of FSL in audio-based classification, including speech recognition and bioacoustic event detection \cite{b11}. In the medical domain, FSL has been used for disease detection using stethoscope recordings and respiratory sounds \cite{b12}. However, the application of FSL to cough-based classification remains underexplored.

While prior studies have largely focused on image-based tasks, our work adapts Prototypical Networks for spectrogram-based cough classification. By using ResNet18 as the backbone and customizing its input layer for single-channel Mel-spectrograms, we demonstrate how vision-based few-shot models can be repurposed for audio classification tasks.

Studies have shown that episodic training strategies improve the generalization of FSL models in medical tasks \cite{b13}. Additionally, hybrid approaches combining transfer learning with few-shot learning have demonstrated promising results in improving classification performance under low-data conditions \cite{b14}. Our work builds upon these advancements by applying Prototypical Networks to the classification of cough sounds and evaluating the impact of transfer learning in few-shot settings. To closely mimic real-world conditions where limited labeled examples are available, we adopt an episodic training framework that aligns with the evaluation strategy of few-shot learning and helps the model generalize better across new classification tasks.

\section{Methodology}

\subsection{Datasets}

This study utilizes three publicly available datasets for cough-based respiratory illness classification: Coswara for healthy coughs, COUGHVID for COVID-19 coughs, and FluSense for Flu coughs. We curated a balanced dataset of 100 samples per class for three classes: Healthy, COVID-19, and Flu. The dataset was split into 80\% training and 20\% testing, ensuring consistency in evaluation across few-shot learning experiments.

\subsubsection{Coswara Dataset (Healthy Coughs)}
The Coswara dataset \cite{aa} was developed by the Indian Institute of Science (IISc) and contains audio recordings of breathing, coughing, and speech samples from individuals who self-reported their health status. The healthy cough subset used in this study consists of recordings from individuals with no reported respiratory conditions.

\subsubsection{COUGHVID Dataset (COVID-19 Coughs)}
The COUGHVID dataset \cite{ab} is a crowdsourced collection of cough recordings developed by EPFL. It contains metadata such as self-reported COVID-19 status, making it useful for training machine learning models for COVID-19 detection.

\subsubsection{FluSense Dataset (Flu Coughs)}
The FluSense dataset \cite{ac} was designed for influenza-like illness surveillance using contactless sensors. It provides real-world Flu cough recordings, making it valuable for differentiating Flu from other types of cough.

\subsection{Preprocessing}
To ensure consistency across datasets, all cough audio recordings were processed as follows:

\begin{itemize}
    \item Duration Normalization: Audio samples were trimmed or zero-padded to 1 second.
    \item Resampling: All recordings were resampled to 22.05 kHz.
    \item Spectrogram Extraction: 128-band Mel spectrograms were extracted using Librosa.
    \item Tensor Conversion: Spectrograms were converted to PyTorch tensors and normalized.
    \item Image Resizing: Spectrograms were resized to 224$\times$224 to match CNN input dimensions.
\end{itemize}

\begin{figure}
    \centering
    \includegraphics[width=0.3\textwidth]{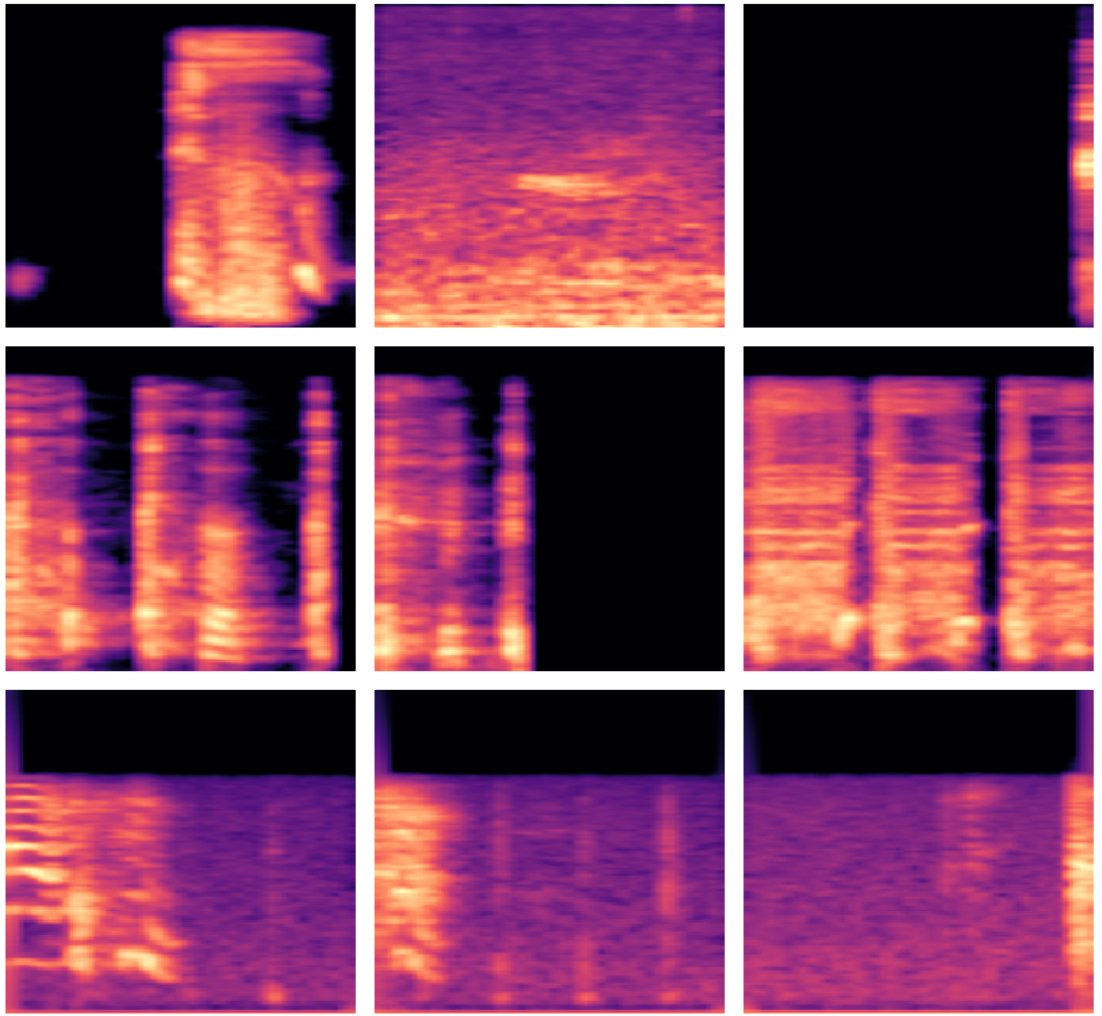}
    \caption{Visualization of support set spectrograms used for training. Top: Healthy, Middle: COVID-19, Bottom: Flu}
    \label{fig:support}
\end{figure}

\begin{figure}
    \centering
    \includegraphics[width=0.3\textwidth]{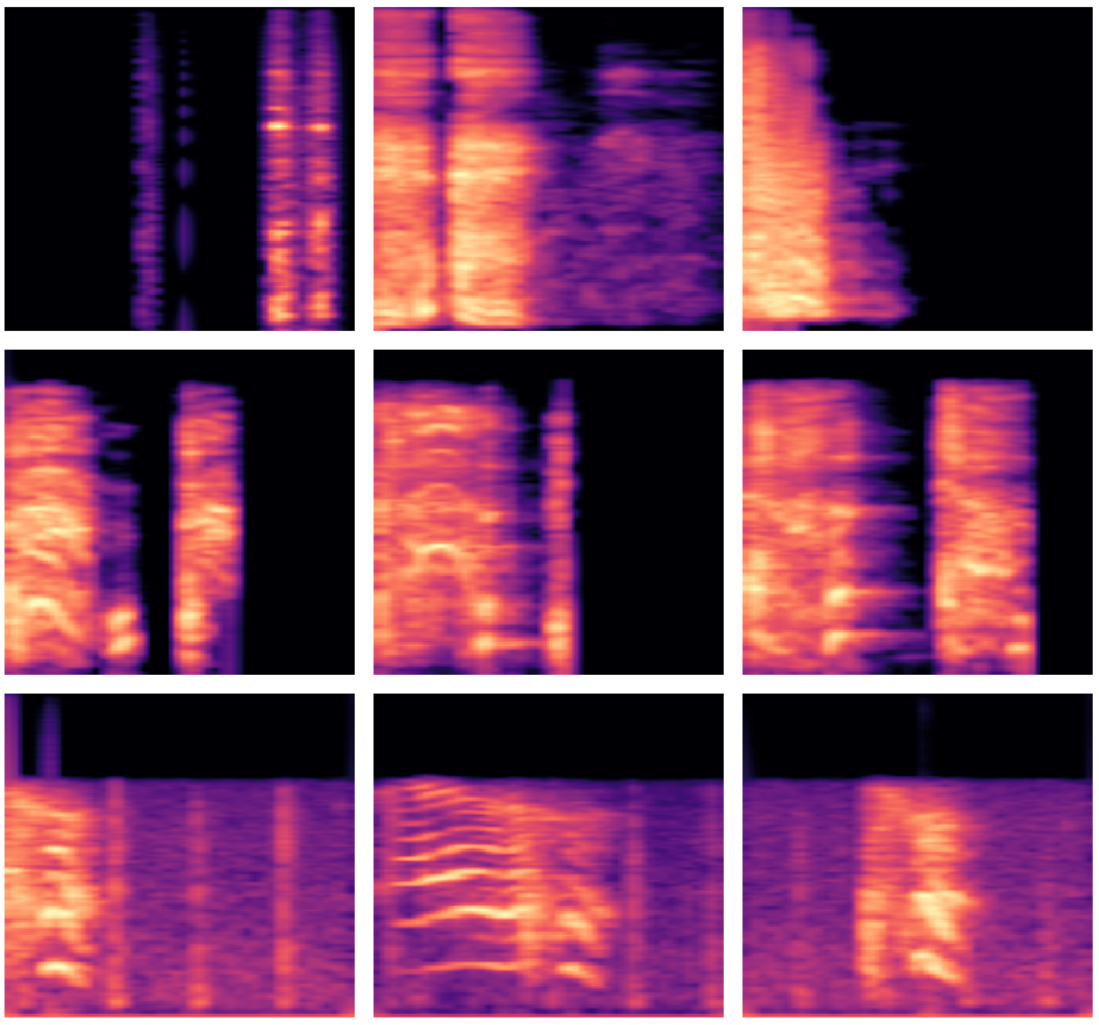}
    \caption{Visualization of query set spectrograms used for evaluation. Top: Healthy, Middle: COVID-19, Bottom: Flu}
    \label{fig:query}
\end{figure}

We visualize spectrograms from support and query sets to demonstrate class separability within few-shot tasks. Each row in Figures~\ref{fig:support} and~\ref{fig:query} corresponds to one class: COVID-19, Flu, and Healthy. These figures highlight how episodic data is organized and fed to the model during training and inference.

\subsection{Model Architecture}

\subsubsection{Prototypical Networks}
\begin{figure}[htbp]
    \centering
    \includegraphics[width=0.45\textwidth]{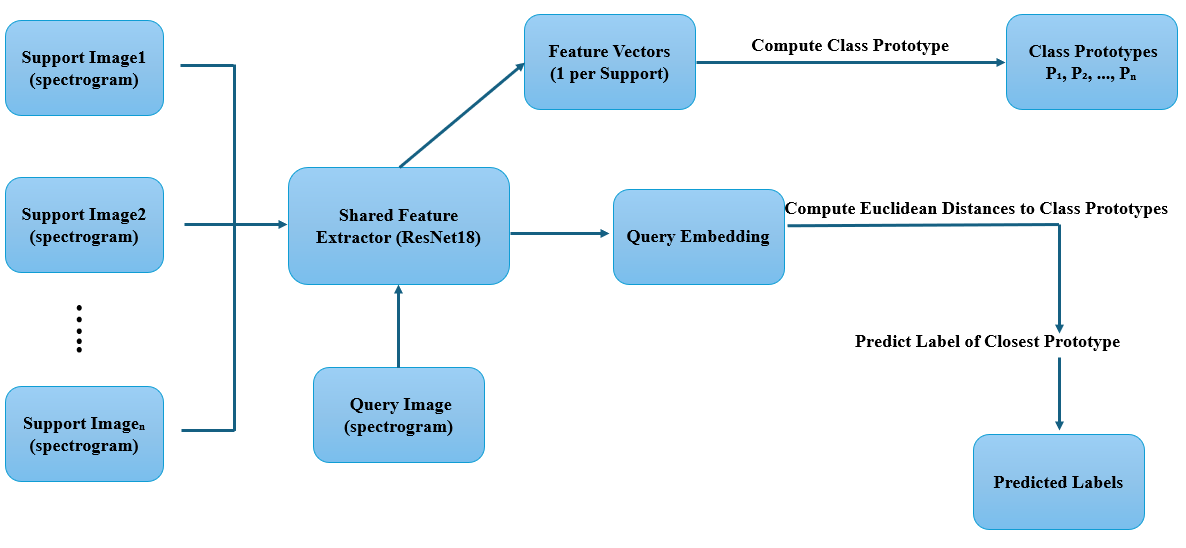}
    \caption{Prototypical Network workflow}
    \label{fig:proto_diagram}
\end{figure}

We use Prototypical Networks \cite{b3}, a metric-based few-shot learning method that computes class prototypes in a learned embedding space. Classification is performed by assigning query samples to the nearest class prototype using Euclidean distance.
Figure~\ref{fig:proto_diagram} illustrates the end-to-end architecture of our Prototypical Network model. Support and query images are passed through a shared ResNet-18 backbone to obtain embeddings. Support embeddings are averaged per class to form prototypes. The query embedding is classified based on its proximity to these prototypes using Euclidean distance.

\subsubsection{Modified ResNet Backbone}
To process 1-channel spectrograms, we modify ResNet-18 \cite{af} as follows:
\begin{itemize}
    \item Replace the first convolutional layer to accept a single input channel instead of 3 (RGB).
    \item Remove the final classification layer and use the model as a feature extractor.
    \item Initialize weights using pretrained ImageNet weights to leverage transfer learning.
\end{itemize}

\subsection{Training and Evaluation Setup}

\subsubsection{Episodic Training Strategy}
The model is trained using an episodic framework:
\begin{itemize}
    \item Support Set: Contains $K$ labeled samples from each of $N$ classes.
    \item Query Set: Contains unlabeled samples from the same $N$ classes.
\end{itemize}
We use the Adam optimizer and tune learning rates to maximize convergence across few-shot tasks. To simulate real-world low-resource classification settings, training and evaluation are performed over multiple episodes (also called few-shot tasks). Each episode randomly samples $N$ classes and selects $K$ support and $Q$ query examples per class. Model performance is averaged over a large number of such episodes (e.g., 100) to ensure stable and generalizable results.

\subsubsection{Evaluation Metrics}
We evaluate the trained model using:
\begin{itemize}
    \item Classification Accuracy: Overall average accuracy for both binary and multi-class tasks.
    \item Class-level Accuracy: Average accuracy breakdown for each class (COVID-19, Flu, Healthy).
    \item t-SNE Visualization: Two-dimensional representation of query embeddings to assess feature separability.
\end{itemize}

\subsection{Experiment Design}

\subsubsection{Experiment 1: Binary Classification Comparisons}
To benchmark performance, we compare binary classification results across three class pairs:
\begin{itemize}
    \item Flu vs. COVID-19
    \item COVID-19 vs. Healthy
    \item Healthy vs. Flu
\end{itemize}

\subsubsection{Experiment 2: Multi-Class Classification}
We evaluate 3-way classification accuracy at different $K$-shot values ($K=1,5,10,15$). This experiment tests the model’s generalization with minimal labeled data.

\subsubsection{Experiment 3: Class-level Accuracy in Multi-Class Tasks}
We track the accuracy of each individual class across $K$ values in multi-class tasks. This identifies whether one class benefits more from few-shot learning.

\subsubsection{Experiment 4: Statistical Analysis}
We conduct two tests to analyze significance:
\begin{itemize}
    \item Paired t-test: Tests the hypothesis that multi-class performance differs significantly from binary.
    \item Wilcoxon Signed-Rank Test: A non-parametric test to validate consistency in relative performance.
\end{itemize}
We compute the absolute difference between binary and multi-class accuracies to determine whether multi-class models fall within a 15\% acceptable range.

\subsubsection{Experiment 5: t-SNE Visualization}
Using the learned embeddings, we apply t-SNE dimensionality reduction to visualize clustering across the three classes. A class-aware scatter plot highlights the ability of the backbone network to separate cough types in the embedding space.

\section{Experimental Findings}

\subsection{Performance of Binary and Multi-Class Classification Models}

The classification performance of the few-shot learning model was evaluated for both multi-class and binary classification tasks using episodic evaluation. Each reported result represents an average over 100 randomly sampled few-shot tasks per $K$-shot setting, providing a stable estimate of generalization across varied support-query combinations.

Table~\ref{tab:binary_accuracy} presents accuracy results for binary classification across varying $K$ values, while Table~\ref{tab:multi_accuracy} summarizes the multi-class performance.

\begin{table}[htbp]
\caption{Binary Classification Average Accuracy (\%) with Standard Error Across K-shot Values}
\begin{center}
\begin{tabular}{|c|c|c|c|}
\hline
K-shot & Healthy vs. Flu & Flu vs. COVID & COVID vs. Healthy \\
\hline
1-shot  & 70.90 $\pm$ 0.50 & 75.10 $\pm$ 0.50 & 61.90 $\pm$ 0.50 \\
5-shot  & 83.80 $\pm$ 3.22 & 89.40 $\pm$ 3.58 & 67.30 $\pm$ 1.35 \\
10-shot & 85.10 $\pm$ 3.20 & 91.50 $\pm$ 3.64 & 70.10 $\pm$ 1.70 \\
15-shot & 88.80 $\pm$ 3.37 & 93.50 $\pm$ 3.62 & 74.80 $\pm$ 1.34 \\
\hline
\end{tabular}
\label{tab:binary_accuracy}
\end{center}
\end{table}

\begin{table}[htbp]
\caption{Multi-Class Model Average Accuracy (\%) with Standard Error Over Different K-Shot Values}
\begin{center}
\begin{tabular}{|c|c|}
\hline
K-shot & Multi-Class Accuracy (\%) \\
\hline
1-shot  & 60.67 $\pm$ 0.50 \\
5-shot  & 68.53 $\pm$ 1.97 \\
10-shot & 71.93 $\pm$ 2.36 \\
15-shot & 72.07 $\pm$ 2.31 \\
\hline
\end{tabular}
\label{tab:multi_accuracy}
\end{center}
\end{table}

As expected, increasing the number of shots improves accuracy across both multi-class and binary classification settings. Binary models consistently outperform their multi-class counterpart, which aligns with previous few-shot learning literature \cite{b11}. Among the binary tasks, Flu vs. COVID-19 achieves the highest accuracy, reaching 93.50\% at $K=15$. The Healthy vs. Flu task also demonstrates strong performance, improving from 70.90\% to 88.80\% across increasing $K$. In contrast, COVID-19 vs. Healthy remains the most challenging pair, reaching 74.80\% at $K=15$.

\begin{figure}
    \centering
    \includegraphics[width=0.45\textwidth]{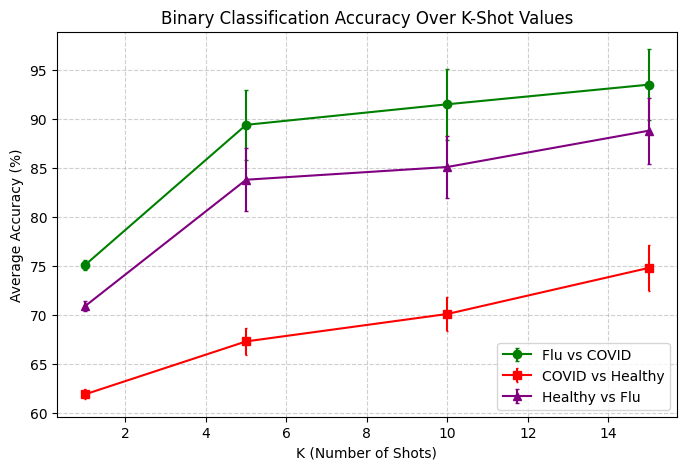}
    \caption{Binary Classification Accuracy}
    \label{fig:binary}
\end{figure}

\begin{figure}
    \centering
    \includegraphics[width=0.45\textwidth]{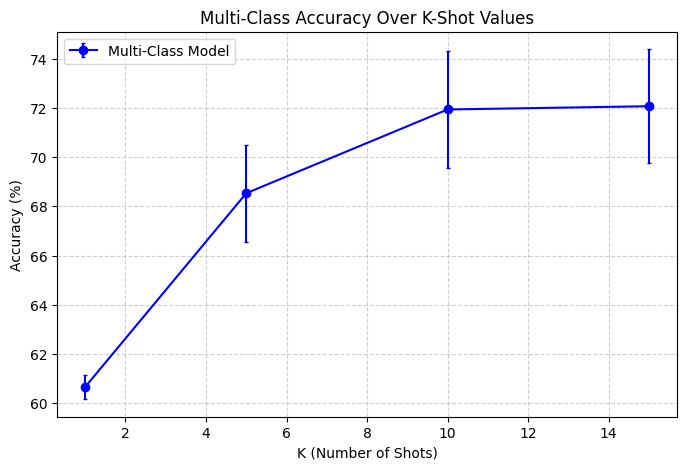}
    \caption{Multi-Class Accuracy}
    \label{fig:multi}
\end{figure}

Figures~\ref{fig:binary} and~\ref{fig:multi} illustrate these trends. In Figure~\ref{fig:binary}, Flu vs. COVID-19 classification shows the steepest improvement, while COVID-19 vs. Healthy lags, possibly due to shared cough characteristics. In Figure~\ref{fig:multi}, accuracy consistently improves with $K$, but the gains diminish after $K=10$, indicating that the model starts to saturate.

\subsection{Multi-Class Accuracy per Class}

To further analyze the multi-class model’s performance, we evaluate class-wise accuracy across different $K$ values. Table~\ref{tab:perclass} summarizes this breakdown.

\begin{table}[htbp]
\caption{Class-level Average Accuracy (\%) of Multi-Class Across Different K-Shot Values with Standard Error}
\begin{center}
\begin{tabular}{|c|c|c|c|}
\hline
\textbf{K-shot} & \textbf{Healthy} & \textbf{COVID} & \textbf{Flu} \\
\hline
1-shot  & 35.0 $\pm$ 0.5  & 64.4 $\pm$ 0.9  & 79.4 $\pm$ 1.0 \\
5-shot  & 56.4 $\pm$ 1.3  & 80.2 $\pm$ 1.3  & 90.8 $\pm$ 1.1 \\
10-shot & 59.8 $\pm$ 1.4  & 86.6 $\pm$ 1.5  & 95.2 $\pm$ 1.5 \\
15-shot & 62.6 $\pm$ 1.4  & 86.2 $\pm$ 1.6  & 96.8 $\pm$ 1.5 \\
\hline
\end{tabular}
\label{tab:perclass}
\end{center}
\end{table}

The Flu class consistently achieves the highest accuracy, suggesting its cough features are the most distinctive. In contrast, the Healthy class is the hardest to classify, possibly due to its broader variability and lack of pathological patterns. The COVID-19 class shows strong accuracy but plateaus slightly beyond $K=10$, indicating some spectral overlap with other classes.

\begin{figure}
    \centering
    \includegraphics[width=0.45\textwidth]{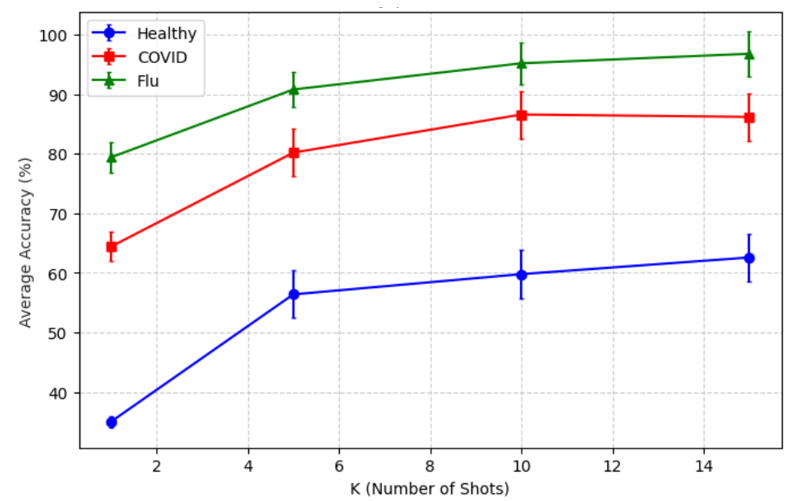}
    \caption{Class-level Average Accuracy (\%) of Multi-Class}
    \label{fig:perclass}
\end{figure}

\begin{figure}[H]
    \centering
    \includegraphics[width=1\linewidth]{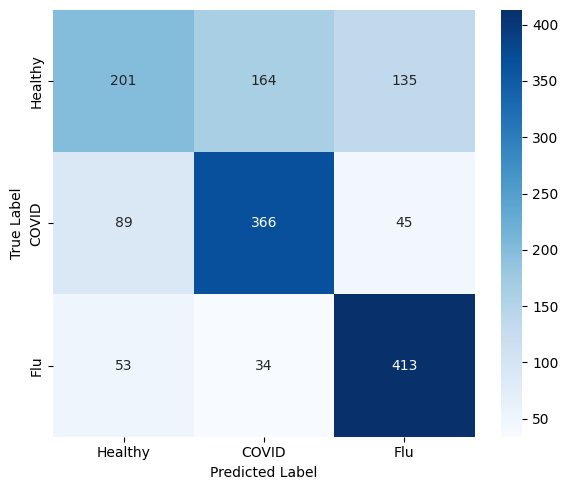}
    \caption{Confusion Matrix for 100 Few-Shot Classification Tasks (3-Way, 5-Shot, 5-Query)}
    \label{fig:confusionmatrix}
\end{figure}

\subsection{Statistical Significance}

We aim to assess whether multi-class classification models—trained using few-shot learning techniques—achieve performance comparable to binary classification models within an acceptable margin of accuracy. Rather than testing for a significant difference, we adopt an equivalence testing framework that evaluates whether the performance gap lies within a practically tolerable threshold.

\subsubsection*{(a) Parametric TOST Equivalence Test}

A Two One-Sided Tests (TOST) equivalence test is used to determine whether two means are statistically equivalent within a predefined margin. Unlike a traditional t-test—which tests for significant differences—TOST assesses whether the observed difference is small enough to be considered practically negligible. This is particularly relevant when the goal is to demonstrate that two models perform similarly enough for real-world deployment, rather than proving one is superior.

To apply this test, we aggregated episodic task accuracy values across all $K$-shot settings ($K = 1, 5, 10, 15$), resulting in 400 episodes for the multi-class model and 1200 for the binary model (averaged over three binary class pairs). We then computed the mean difference in accuracy and a 90\% confidence interval (CI) around that difference using the pooled standard error and $t$-distribution. If this CI lies entirely within the predefined equivalence margin, the null hypothesis is rejected, indicating statistical equivalence.

\begin{table}[H]
\centering
\caption{Equivalence Test Between Multi-Class and Binary Models (TOST)}
\label{tab:equivalence_tost}
\begin{tabular}{|l|c|}
\hline
\textbf{Metric} & \textbf{Value} \\
\hline
Mean Average Accuracy (Multi-Class) & 73.22\% \\
Mean Average Accuracy (Binary)      & 79.66\% \\
Mean Difference             & $-6.44$\% \\
90\% Confidence Interval    & [$-7.77$\%, $-5.11$\%] \\
Equivalence Margin          & $\pm15$\% \\
Result                      & Statistically Equivalent \\
\hline
\end{tabular}
\end{table}

Since the confidence interval falls entirely within the equivalence bounds of $\pm15\%$, we conclude that the performance of the multi-class model is statistically equivalent to that of the binary models. This supports our central research hypothesis that multi-class few-shot models can perform comparably to binary models in diagnostic tasks, with differences remaining within a 10--15\% margin.

\subsubsection*{(b) Non-Parametric Bootstrap-Based Equivalence Test}

To verify that the equivalence conclusion is robust to distributional assumptions (such as normality), we performed a non-parametric alternative to the TOST procedure using bootstrapping. This involved resampling the multi-class and binary accuracy values 10{,}000 times, computing the difference in mean accuracy for each iteration. The 90\% confidence interval was then obtained from the empirical distribution of these differences.

\begin{table}[H]
\centering
\caption{Bootstrap-Based Equivalence Test}
\label{tab:equivalence_bootstrap}
\begin{tabular}{|l|c|}
\hline
\textbf{Metric} & \textbf{Value} \\
\hline
Bootstrap 90\% CI        & [$-7.78$\%, $-5.12$\%] \\
Equivalence Margin       & $\pm15$\% \\
Result                   & Statistically Equivalent \\
\hline
\end{tabular}
\end{table}

The bootstrap-based confidence interval closely mirrors the parametric result, further confirming that the observed performance gap is both statistically and practically negligible. Since the entire interval lies within the pre-specified equivalence bounds, we conclude that the multi-class and binary few-shot models are statistically equivalent—regardless of distributional assumptions.

\subsection{Feature Representation Analysis}

To analyze how the model internally separates classes, we visualized the learned embeddings using t-SNE.

\begin{figure}[H]
    \centering
    \includegraphics[width=0.45\textwidth]{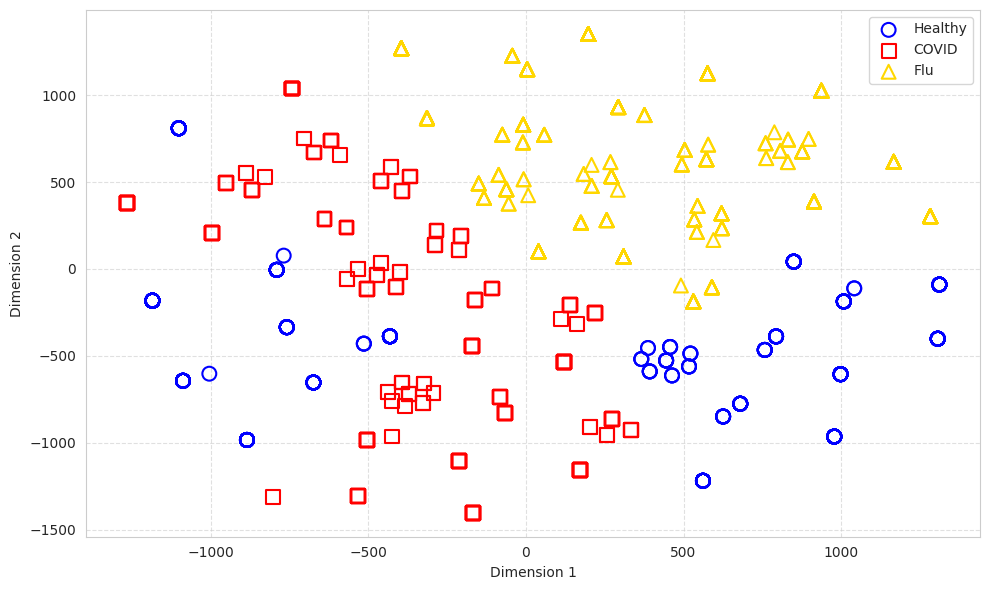}
    \caption{t-SNE Visualization of Data Distribution In-terms of Feature Representations}
    \label{fig:tsne}
\end{figure}

As shown in Figure~\ref{fig:tsne}, Flu samples form a well-defined cluster, whereas Healthy and COVID-19 samples show overlap. This visualization aligns with the observed class-level accuracy trends and reinforces the notion that Flu coughs exhibit more distinguishable acoustic patterns.

\subsection{Comparison with Existing Studies}

Compared to prior deep learning models trained on large-scale datasets, our few-shot model achieves comparable results with significantly fewer labeled examples. While conventional CNN-based studies report high performance \cite{am}, they typically require thousands of samples to generalize well. \cite{b6} reported a high accuracy on crowdsourced respiratory sound data by leveraging a large dataset and extensive training. While these results are impressive, they come at the cost of requiring extensive data collection, annotation, and computational resources. In contrast, our model, trained on only 100 samples per class, demonstrates the ability of few-shot learning to deliver competitive performance in data-constrained healthcare scenarios.

This approach not only reduces the annotation burden but also enables faster deployment in emerging diagnostic settings, where labeled data is often scarce or unavailable.

Recent literature reinforces the promise of few-shot learning in medical domains. A systematic review by Suri et al. highlights that few-shot techniques can significantly address data scarcity challenges in medical imaging while maintaining robustness and interpretability \cite{new1}. Islam et al. surveyed clinical applications of data-efficient learning and reported that few-shot methods achieve promising accuracy across imaging modalities \cite{new2}. Yan et al. introduced episodic-based strategies to enhance generalization in few-shot classification tasks under real-world data constraints \cite{new3}. In another study, Zhang et al. proposed a class-prototype refinement approach that improved few-shot tumor recognition using limited examples \cite{new4}. Similarly, Rahman et al. applied few-shot techniques for COVID-19 chest X-ray classification and showed that they can match large-scale baselines with far fewer training samples \cite{new5}.

These findings support our experimental results and affirm that few-shot learning is a practical and scalable solution for healthcare tasks where annotated data is limited, especially in dynamic or resource-constrained environments.

\section{Discussions}



Despite encouraging results, our study has several limitations. First, the performance in extreme low-shot settings (e.g., \( K = 1 \)) remains limited, suggesting that there is room for improving the current embedding strategies. Second, the datasets used in this work rely on self-reported labels, which may introduce noise or labeling inaccuracies that could affect model performance. Finally, healthy coughs remain the most challenging class to identify, likely due to their greater acoustic variability and the absence of consistent, structured patterns typically found in pathological coughs. Additionally, the healthy samples in this study were sourced from the Coswara dataset, a COVID-focused collection in which participants self-reported their health status. This may have introduced further ambiguity, as some healthy-labeled samples could acoustically resemble COVID-19 coughs due to their origin from a COVID screening context and the potential overlap in self-induced labeling practices.

While Few-Shot Learning (FSL) techniques are designed to reduce data requirements, they can be computationally demanding. Despite using a relatively lightweight backbone like ResNet-18, our experiments involved extensive episodic training.
We conducted all training and evaluation on Google Colab with GPU acceleration, and the complete training process took approximately 6 hours. This highlights that, contrary to common assumptions, FSL approaches—especially those using episodic frameworks and convolutional backbones—may still require moderate to high computational resources for convergence and generalization.

Future directions include exploring self-supervised pretraining to enhance feature robustness and combining Prototypical Networks with attention mechanisms for improved discrimination in overlapping class scenarios. 
Alternative FSL strategies to Prototypical Networks include Matching Networks, which leverage an attention mechanism for one-shot learning; Relation Networks, which learn a deep distance metric between embeddings; and Model-Agnostic Meta-Learning (MAML), which trains an adaptable initialization for rapid learning across tasks. While Prototypical Networks offer simplicity and efficiency, future studies could investigate whether MAML-based or transformer-backed meta-learners yield improved generalization for cough classification.
Additionally, expanding to larger and more clinically verified datasets could improve generalizability and real-world reliability.

\end{document}